\pdfoutput = 1
\documentclass[letterpaper, 10pt, conference]{ieeeconf}

\usepackage{etoolbox}
\makeatletter
\patchcmd{\@makecaption}
  {\scshape}
  {}
  {}
  {}
\patchcmd{\@makecaption}
  {\\}
  {.\ }
  {}
  {}
\patchcmd{\@makecaption}
  {\centering}
  {}
  {}
  {}
\makeatother

\IEEEoverridecommandlockouts  
\makeatletter
\let\NAT@parse\undefined
\makeatother

\usepackage{cite} 
\usepackage[pdftex]{graphics} 
\usepackage{epsfig} 
\usepackage[cmex10]{amsmath} 
\usepackage{newtxmath} 
\usepackage{times} 
\usepackage[hidelinks]{hyperref} 
\usepackage[all]{hypcap} 
\usepackage[capitalise,nameinlink]{cleveref} 
\usepackage{algorithm}
\usepackage{algpseudocode}
\usepackage{bm}
\usepackage{multirow}
\usepackage[subrefformat=parens,labelformat=parens,caption=false,font=footnotesize]{subfig}
\usepackage{array}
\usepackage{pifont}
\usepackage{balance}
\usepackage[flushleft]{threeparttable}
\usepackage{todonotes}

\usepackage{booktabs}

\newcommand\norm[1]{\left\lVert#1\right\rVert}

\title{\LARGE \bf
Modular Neural Network Policies for Learning In-flight Object Catching with a Robot Hand-Arm System
}

\author{Wenbin Hu$^{1}$, Fernando Acero$^{2}$, Eleftherios Triantafyllidis$^{1}$, Zhaocheng Liu$^{1}$, Zhibin Li$^{2}$
\thanks{$^{1}$School of Informatics, University of Edinburgh, UK.}
\thanks{$^{2}$Department of Computer Science, University College London, UK}
\thanks{F.A. and E.T. are supported by the EPSRC Foundational AI CDT and EPSRC CDT in Robotics and Autonomous Systems respectively.}
}

\begin{document}

\maketitle

\begin{abstract}
We present a modular framework designed to enable a robot hand-arm system to learn how to catch flying objects, a task that requires fast, reactive, and accurately-timed robot motions. Our framework consists of five core modules: (i) an object state estimator that learns object trajectory prediction, (ii) a catching pose quality network that learns to score and rank object poses for catching, (iii) a reaching control policy trained to move the robot hand to pre-catch poses, (iv) a grasping control policy trained to perform soft catching motions for safe and robust grasping, and (v) a gating network trained to synthesize the actions given by the reaching and grasping policy. The former two modules are trained via supervised learning and the latter three use deep reinforcement learning in a simulated environment. We conduct extensive evaluations of our framework in simulation for each module and the integrated system, to demonstrate high success rates of in-flight catching and robustness to perturbations and sensory noise. Whilst only simple cylindrical and spherical objects are used for training, the integrated system shows successful generalization to a variety of household objects that are not used in training.  
\end{abstract}

\section{Introduction}\label{seq:introduction}
\begin{figure}
    \centering
    \includegraphics[trim= 180 105 180 125, clip, width=\linewidth]{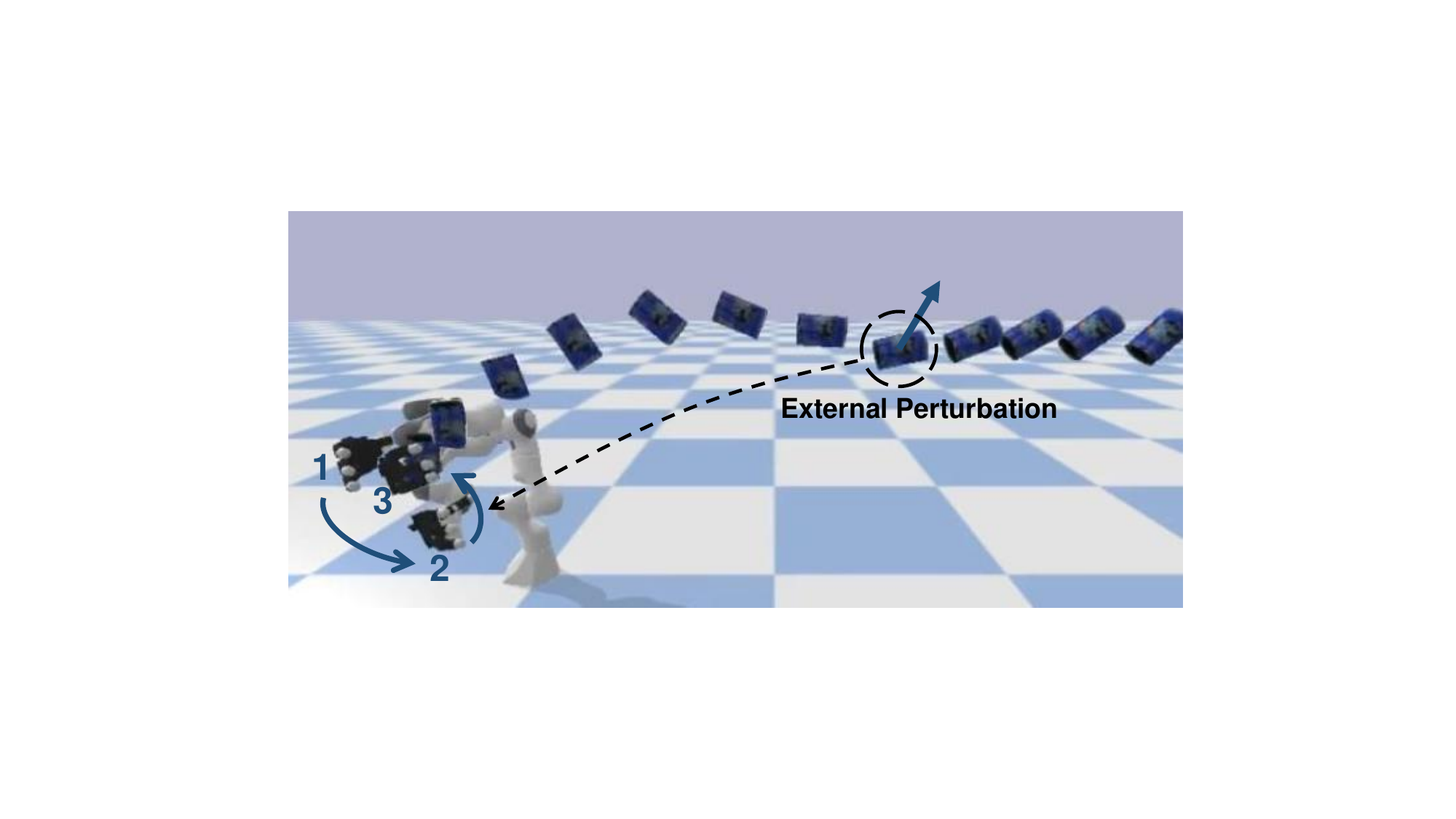}
    \vspace*{-9mm}
    \caption{Catching a flying object under an unexpected external perturbation: labels 1-2-3 show reactive motions of the robot hand; the robot constantly re-estimates the flight trajectory, adjusts the motion rapidly, and catches the object successfully.}
    \label{fig:first_page_figure}
    \vspace{-6mm}
\end{figure}

Humans are capable of interacting with flying objects in a variety of scenarios ranging from ball sports to brick-tossing in construction works. In contrast, the current capability of robot manipulation is largely restricted to industrial environments, and dynamic tasks with high variability such as catching flying objects still remain challenging.
To catch a flying object, within an extreme short duration (usually less than 1 second), the robot has to accomplish a sequence of sub-tasks including: accurate object trajectory prediction, catching pose determination and real-time motion generation. In this work we propose an end-to-end learning framework to address the problem of catching flying objects with a multi-joint robot hand-arm system.
As shown in \cref{fig:first_page_figure}, for a flying object under a sudden perturbation, our system quickly re-estimates the object trajectory, adapts the pre-catch pose, and successfully catches the object at a new pose.

Approaches to solving the problem of catching flying objects can be broadly classified in two categories: non-prehensile catching \cite{2021_xiongrong, 2020_KeDong} and prehensile catching \cite{2010_ICIRS, 2011_humanoid, 2014_catching, 2016_soft_catch}.
Most non-prehensile catch approaches focus on accurate prediction of the object trajectory, where the object is caught by the non-prehensile end-effector, e.g. a net or a cup.

For prehensile catching, the required robot motion is more demanding. Firstly, the determination of the catching pose should consider both the morphology and motion of the object. For example, a dexterous hand ought to grasp cylindrical objects from lateral directions, and the object velocity is better orthogonal with the palm instead of being parallel with it. The previous method uses human demonstrations to determine the pre-catch pose of target objects, neglecting the object motion when the catching happens \cite{2014_catching}. While in this paper we proposed a catching pose quality network to quantify the feasibility of the hand-object pose tuple for catching, considering both the object velocity and the required robot motion.
Secondly, the motion of the hand and arm need to be coordinated. To extend the grasping duration and ease the required precision of grasping timing, a prescribed soft catching strategy has been proposed \cite{2016_soft_catch}, where the robot arm moves with the object for a short period of time. In this work, the self-emergent soft catching policy is learned via deep reinforcement learning (DRL).

Since the final catching configuration frequently changes as the object trajectory prediction updates, the robot movements need to be highly dynamic and responsive.
It is difficult to re-plan the robot trajectory using nonlinear optimization approaches with a high frequency online \cite{2011_humanoid}.
Encoding the arm and hand as Dynamical Systems (DS) \cite{2011_DS} is one approach to catching tasks \cite{2016_soft_catch, 2014_catching}, where kinematically feasible trajectories of the robot arm and hand are collected as demonstrations.

Alternatively, DRL is a different approach to learn manipulation skills \cite{2019_contactforce, 2021_wenbin} without the dependence on human demonstrations. In this work, we train the motion generation policies with DRL: the trained policies output control commands based on current state observations. The state-action control scheme enables the policies to generate highly dynamic robot motion and react rapidly to environmental variations and sensory uncertainties.
Here we propose a systematic framework for catching flying objects, consisting of five different modules, as numerated in \cref{fig:system_overview}: (i) an \textbf{object state predictor} for estimating the flying object trajectory; (ii) a \textbf{catching pose quality network} for choosing the best catch configuration; (iii) a \textbf{reaching policy} for moving the robot hand to the selected pre-catch pose; (iv) a \textbf{grasping policy} for performing a soft catching motion to reduce the contact impact between hand and object; and (v) a \textbf{gating network} to coordinate the reaching and grasping policies.
Though the modules are designed for catching flying objects, they can easily be deployed to other tasks with minor modifications, e.g. grasping objects on a moving conveyor.
The contributions of our work are as follows: 
\begin{itemize}
\item[(1)] A catching pose quality network to evaluate and select the catching poses, which considers both the quality of the target object pose and the level of difficulty for the hand to approach it.
\item[(2)] A gating network to synthesize both the robot hand and arm motions, to seamlessly and smoothly coordinate the reaching and soft catching of in-flight objects.
\item[(3)] An integrated framework with five learning-based modules to catch flying objects of various shapes and be robust in presence of sensory noise and perturbations.
\end{itemize}

In the following sections, we first review the literature related to robot catching in \cref{seq:related_work}. Our methodology is elaborated in \cref{seq:methods}, and our proposed system is evaluated in \cref{seq:validation}. Finally, we summarize our work and propose future research directions in \cref{seq:conclusion}.
\section{Related Work}\label{seq:related_work}
\subsection{Object trajectory prediction}
Accurate motion prediction of in-flight objects is crucial for catching tasks.
A ball is the most frequent target object in robot catching tasks, because its trajectory is relatively easy to predict.
When the ball is small and has uniform mass distribution, the flying trajectory can be approximated as a parabola \cite{2017_ChenICAR}.
Apart from gravitational forces, aerodynamic drag is the most significant factor that needs to be considered in trajectory prediction. However, the air drag coefficient typically requires experimental calibration and is related to the object's shape \cite{2011_humanoid}. 
One mitigation strategy to the above problem is the use of the Extended Kalman Filter (EKF), which is a widely used estimator for the object's state and allows the consideration of air drag and other external factors \cite{2010_ICIRS, 2020_KeDong}. 
Developing explicit models of the object dynamics is another option for trajectory prediction \cite{2019_battingIJRR}. However, requirements on prior knowledge such as mass or moment of inertia limit the generalization ability.

Estimating the dynamics model with machine learning methods has achieved promising results \cite{2012_kim_estimation}, but the learned models suffer from performance drop when generalizing to novel objects due to limited data size.
Learning-based models are widely used to approximate nonlinear dynamics. In \cite{2021_xiongrong} a neural network-based model and a differentiable Kalman filter are trained to estimate acceleration of an uneven object by observing the previous detected trajectory. 
In this work, we learn the object trajectory prediction using a recurrent neural network (RNN) with access to a short time history of the object trajectory.

\subsection{Catching pose selection}
When the object trajectory is predicted, a feasible catching pose intersecting the object trajectory and within the workspace of the robot arm should be selected.
Moreover, the hand should be able to reach the selected target pose before the object arrives.
For catching spherical objects, a common way to determine the interception pose is to find the nearest intersection between the object's flying trajectory and the robot's reachable space.
The pose selection can be formulated as an optimization problem with nonlinear constraints \cite{2010_ICIRS}, which can be solved by quadratic programming.
For objects without central symmetry, the robot hand has to attain certain orientation before proceeding to catch. In \cite{2014_catching}, researchers used a trained graspable space model of the specific target object to predict the hand pose.
The aforementioned algorithms neglect the object motion when considering the pre-catch pose of robot hand.
We propose a neural network based scoring model to evaluate the candidate object poses by the control effort required for the robot to move towards them from the current joint configuration, as well as the object velocity direction relative to the hand.

\begin{figure}[t]
    \centering
    \includegraphics[trim=190 10 190 10, clip, width=\linewidth]{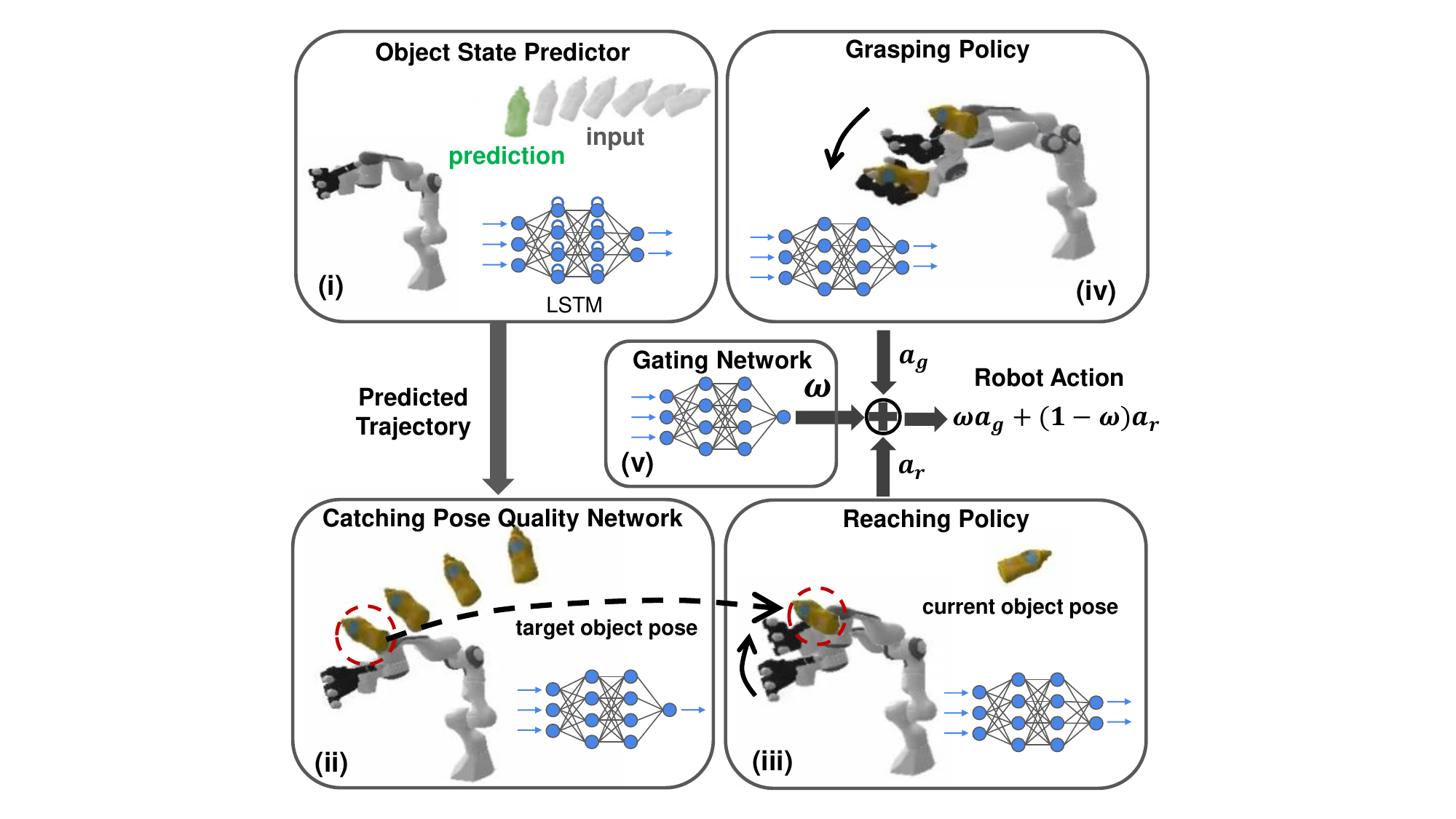}
    \vspace*{-5mm}
    \caption{Integrated framework consisting of five neural network modules to learn to catch flying objects.}
    \label{fig:system_overview}
    \vspace{-5mm}
\end{figure}

\subsection{Robot motion generation}
Given a selected catching pose, to enable the robot to intercept the flying object at a particular time with a particular posture, in \cite{2014_catching} the authors used time-invariant dynamic systems to encode the robot's motion. 
Learning from demonstrations \cite{2017_hitting} is also possible, where the motion dynamics can be modeled from the expert demonstrations, e.g. kinesthetic teaching.
However, the generalization ability is limited by the scale and nature of the demonstration set.
DRL offers an alternative approach for learning robot control \cite{2020_dronecatch}. Unlike aforementioned algorithms where the time-variant robot motion is planned in advance, we propose a DRL scheme, where the learned control action is generated based on the current state observation to maximize expected future reward. This state-action based control approach can react to the environmental changes rapidly, which is crucial for object catching tasks.

\section{Methods} \label{seq:methods}

Our proposed system consists of five distinct modules, as illustrated in \cref{fig:system_overview}. 
The object state predictor estimates the flying trajectory, and the catching pose quality network evaluates the poses on the trajectory and selects the desired object pose when the catching motion happens. These two modules have strict demand of accuracy and hence we model them as neural networks and train them with supervised learning using large amount of synthetic data.
The remaining three modules control the robot and they are trained with deep reinforcement learning, because the state-action control manner is effective for the highly dynamic tasks.
The reaching policy network moves the robot hand to the corresponding pre-catch pose and then grasping policy network accomplishes the grasping motion with mitigated impact forces.
The gating network coordinates the reaching and grasping policies by blending the output actions of both policies at real time.

The first two modules work at 100Hz, whereas the last three modules work at 50Hz.
The modules are trained and validated using PyBullet physics simulator \cite{pybullet}.
Based on the dependencies between the five modules, the order to train is: (1st) object state prediction network and reaching policy (separately); (2nd) catching pose quality network; (3rd) grasping policy; (4th) gating network.

\subsection{Object Trajectory Prediction} \label{subseq:predictor}
An accurate prediction of the future motion trajectory is relevant to catching performance.
Explicitly modeling the nonlinear dynamics system is difficult, so a practical approach is to estimate the object velocity and acceleration with a trained neural network. The Long Short-Term Memory (LSTM) network is widely used in processing and extracting intrinsic information from sequential data \cite{HochSchm97, 2021_xiongrong}.
We use an LSTM network which takes a sequence of past object states with $n$ time-steps $[X^o_{t-n+1}:X^o_{t}]$ as input, to predict the object state at next time-step $X^o_{t+1}$. 
The network has one LSTM cell with 100 features in the hidden state, followed by one fully connected hidden layer with 100 neurons.

The future object trajectory is then derived from the predicted object state.
The object state vector $X^o$ consists of positions, orientations, linear/angular velocities and linear/angular accelerations, where the velocities and accelerations are computed by numerical differential. To simulate air resistance in a simple manner, linear damping is introduced on the linear and angular velocity of the objects, which aerodynamically corresponds to a Stoke's drag assumption. The training dataset is gathered in simulation with simple objects (see \cref{subseq:module_validation}).
Since the object trajectory prediction model is trained separately to other modules, for real-world experiments, we can replace the simulated training data with real object flight trajectories and use a more complex neural network structure, such as the one proposed in \cite{2021_xiongrong}.

\subsection{Catching Pose Quality Network} \label{subseq:scorer}

Given the current pose of robot hand, the success of a catching attempt is highly dependent on the choice of the catching timing, or equivalently, the choice of the object pose when the grasping action is performed.
The proposed catching pose quality network is used to evaluate the capability of the robot hand at current pose $P^h$ to successfully catch the object at pose $P^o$. 
The network takes both current robot hand pose and candidate object pose as input: $[p^o, q^o, p^h, q^h]$, where $p$ denotes Cartesian position and $q$ denotes the orientation quaternion, and outputs a scalar quality score which quantifies the effectiveness of the hand to reach and grasp the object.
The network has 2 fully-connected hidden layers, each with 100 neurons.

With the current hand pose, the trained network generates the scores of all the object poses on the predicted flight trajectory, and then the one with highest score will be selected as the target object pose when the catching happens, and the robot hand ought to reach the corresponding pre-catch hand pose before the object arrives.
To gather the training data, we fixed the object at a random pose and the robot is controlled by the trained reaching policy to reach the pre-catch pose, as described in \ref{subseq:reaching_policy}. During the motion, the pre-catch pose and the robot arm joint positions $J$ are recorded.
The computation of the score consists of two parts: the required time and movements for the hand to reach the pre-catch pose, and the effectiveness of the pre-catch pose in aiding the catching of the incoming object.
The score is defined as:
\begin{equation}
    s = e^{-\norm{J_d-J_{now}}} \cdot e^{-\norm{p_d^h-p^o}} \cdot \left(1 -\vec{u}_{n} \cdot \frac{\vec{v}}{\vert \vert \vec{v} \vert \vert}\right),
\end{equation}
where $J$ denotes the vector of robot arm joint positions; $\vec{u}_n$ denotes the normal unit vector of the palm, pointing outwards; $p$ denotes the position and $\vec{v}$ denotes the object velocity vector. 
The subscript $d$ denotes the desired pre-catch pose of the hand.
The first term evaluates the changes in the robot arm joint configuration during the reaching motion, indicating the required time and efforts for moving the hand from current pose to the desired pre-catch pose. 
The second term evaluates the distance between the pre-catch hand position $p_d^h$ and the object position $p^o$.
The third term relates to the orientation of the palm when it contacts the object. To increase the contact area during the catching motion, the plane of the palm should be perpendicular to the object's moving direction. The constant value $1$ is added to this term, shifting the value range from $[-1, 1]$ to $[0, 2]$.
The second and third term evaluate the effectiveness of the pre-catch hand pose in catching the moving object at the target pose.
The object pose in flying trajectory which is most adequate for catching can be selected by the network. However, if the object flying trajectory is either too far from the robot hand or in an inappropriate pose, the object pose with highest score can be unfeasible for catching.

The catching pose quality network is trained in a supervised manner. 
The data gathering procedure consists of two steps: Firstly, $N$ object flying trajectories $\tau_{1:N}$ with random initial object poses and velocities are sampled and logged. Then, for every trajectory, $M$ object poses $P^o_{1:M}$ with fixed time interval are selected as the target object pose. 
$K$ random hand poses $P^h_{1:K}$ within a certain range are also sampled and logged.
Secondly, given the object fixed at one selected target pose $P^o_i$, the robot hand will reach it from the selected hand pose $P^h_j$, controlled by the learned reaching policy. After a fixed period of time, the corresponding score for the hand-object pose tuple $[P^o_i, P^h_j]$ is computed and logged, and the current robot hand pose is regarded as the desired pre-catch pose.
In the training data gathering procedure, $N*M*K$ data points are collected.

\begin{figure}
    \centering
    \includegraphics[trim=250 120 250 120, clip, width=0.7\linewidth]{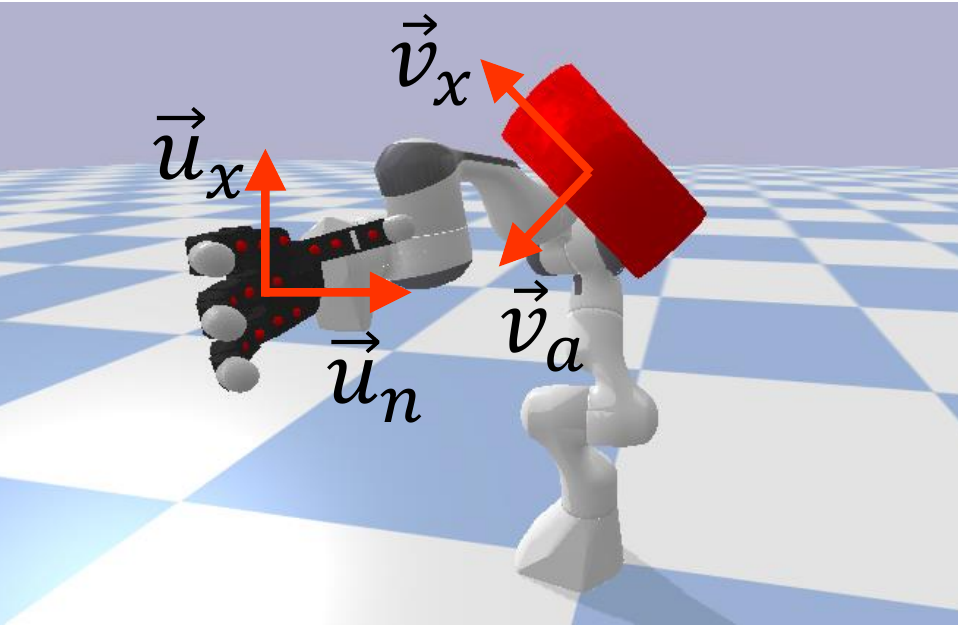}
    \caption{Illustration of the assignment of vectors and hand key-points for the reward function.}
    \label{fig:vectors}
    \vspace{-5mm}    
\end{figure}

\subsection{Reaching Policy Network} \label{subseq:reaching_policy}
Given the target object pose selected by the catching pose quality network, the reaching policy will determine a corresponding pre-catch hand pose and control the robot to reach it as soon as possible.
The reaching policy is trained with DRL, in a trail-and-error manner, where the robot gradually updates the policy by gathering rewards through interaction with the environment.
The learning algorithm is Proximal Policy Optimization (PPO), a widely used DRL algorithm for continuous control tasks \cite{PPO}.
To train the policy, the object is fixed at a random pose throughout the episode. To stabilize the physics simulator and accelerate learning, the contact check between the object and the robot is disabled.

The state space is $S_r = \{\Tilde{p}, \Tilde{q}, \vec{v}_a \}$, where $\Tilde{p}$ and $\Tilde{q}$ denote the position and quaternion of the selected target object pose relative to the robot hand, and $\vec{v}_a$ denotes the unit vector showing the desired direction for the hand to approach, pointing from the object to the outward, as shown in \cref{fig:vectors}.
The action space $A_r$ consists of the linear velocity and angular velocity of the robot hand. The robot arm joint positions are computed by inverse kinematics given the desired end-effector pose.
When the robot is controlled by the reaching policy, the finger joints maintain an open configuration, prepared for catching the object.

The reward function of the reaching policy is the linear combination of three different terms:
\begin{equation}
    R_r = -\frac{1}{k}\sum_{i=1}^{k}\norm{p^{k_i}-p^o} + \left( -\vec{u_n} \cdot \vec{v}_a \right) + 
    \lvert\vec{u}_x \cdot \vec{v}_x\rvert .
\label{eq:reward_reach}    
\end{equation}
The first term is the negative of mean distance between the target object position $p^o$ and the hand key-points positions $p^{k_i}$. As shown in red color in \cref{fig:vectors}, the $k$ key-points are equally distributed on the inner surface of the robot palm and fingers. This term will reward the hand to approach the object.
In the second reward term, $\vec{u}_n$ denotes the unit normal vector of the palm. $\vec{v}_a$ denotes the approaching vector of the object, which is always orthogonal with the object's major axis.
For the flying object, $\vec{v}_a$ is also coplanar with the gravity vector and the object velocity vector.
For small or spherical objects without major axis, the $\vec{v}_a$ is the unit linear velocity vector.
This term will reward the hand to increase the contact area by adjusting the hand orientation.
The third reward term is the absolute value of the dot product between $\vec{u}_x$ and $\vec{v}_x$ where $\vec{u}_x$ denotes one of the major axis of the hand, and $\vec{v}_x$ denotes the major axis of the cylindrical object. 
This term rewards the hand to align with the object's major axis, similar to the way that humans grasp cylindrical objects.
The four aforementioned vectors are illustrated in \cref{fig:vectors}.
We also add early termination criteria where the robot arm approaches a singularity or the joints hit the position or torque limits, aiming to learn safe robot motion.
The reaching policy network has 2 fully-connected hidden layers, each with 256 neurons, and it is trained with 4 million time steps.

\begin{figure}[t]
    \centering
    \includegraphics[width=\linewidth]{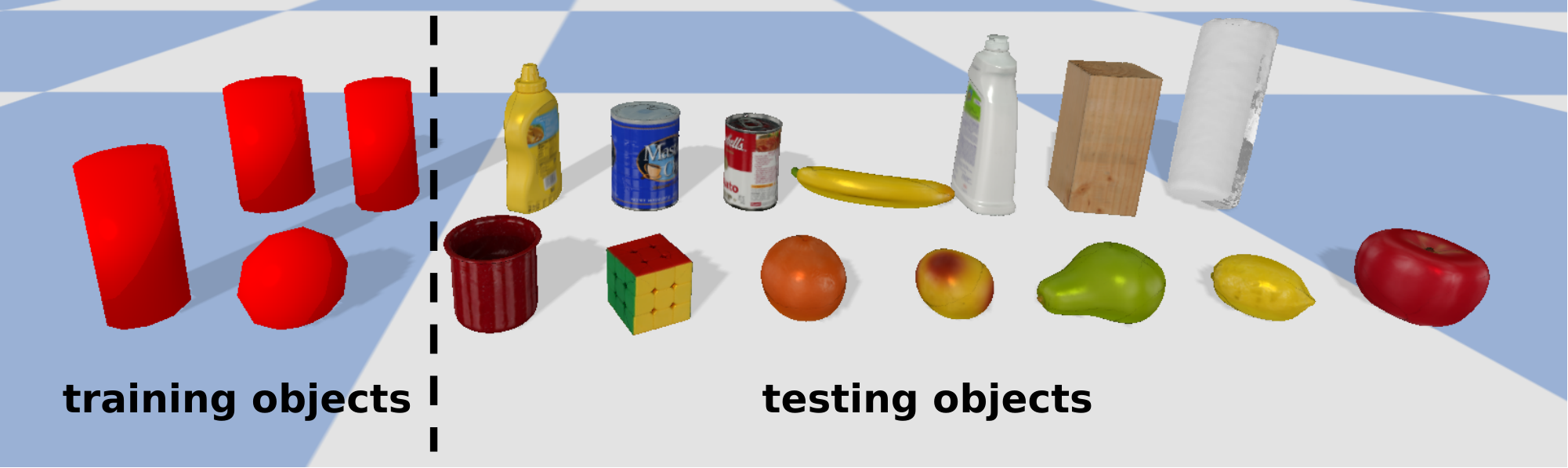}
    \vspace*{-5mm}
    \caption{Objects used for training (red cylinders and spheres), and the various household objects (right section) used to test the generalization of the integrated catching system. }
    \label{fig:objects}
    \vspace{-5mm}    
\end{figure}

\subsection{Grasping Policy Network} \label{subseq:grasping_policy}

The grasping policy is also trained with PPO to perform the final catching motion when the object is approaching.
The learning of the grasping policy requires the trained three aforementioned modules.
In every training episode, the object is tossed from a random pose with a random velocity, flying towards the robot (see \cref{subseq:system_validation} for randomization range). 
The future object trajectory is predicted at 100 Hz. Then the catching pose quality network selects the target object pose, and the reaching policy moves the hand towards the corresponding pre-catch pose.
Once the object is estimated to arrive at the selected target pose after time $T$, the grasping policy takes over the robot control to interact with the environment.
We refer to $T$ as the \textit{preparation time} for the grasping policy. If $T$ is too long, the grasping policy might take over the control too early, before the hand arrives the pre-catch pose, and if $T$ is too short, the hand might not be able to reach the best catching speed and fingers might not be able to close in time.
To increase the robustness of grasping policy, the preparation time is randomized during training: $T \in [0.05, 0.25]$s. The seamless switch between reaching and grasping policies is learned by gating network as described in \ref{subseq:gate_network}.

The state space of the grasping policy $S_g$ consists of the object pose relative to the hand, linear velocity of the object, linear velocity of the hand and the joint positions of the fingers.
The action space of the grasping agent $A_g$ consists of the linear, angular velocities of the hand, and the target finger joint positions.
The reward function used for training the grasping policy is the sum of three different terms:
\begin{equation}
    R_g = -\frac{1}{k}\sum_{i=1}^{k}\norm{p^{k_i}-p^o} + \frac{n_c}{k} + r_p .
\label{eq:reward_grasp}
\end{equation}
The first term of \cref{eq:reward_grasp} is the same as the first term of \cref{eq:reward_reach}, both rewarding the hand to approach the object. However, here we use the real-time object pose, while in \cref{eq:reward_reach}, we use the target object pose selected by the catching pose quality network.
In the second term, $n_c$ denotes the number of the hand key-points that are in contact with the object. This term rewards the hand to grasp the object with larger contact area.
The last term $r_p$ is a piece-wise function defined as:
\begin{equation}
    r_p= \begin{cases} 
      e^{-\norm{p^h-p^o}}\cdot \vec{v}_h \cdot \vec{v}_o & n_c = 0 \\
      e^{-\norm{\vec{v}_h}}& n_c > 0 
      \end{cases}
\label{eq:reward_velvec}      
\end{equation}
where $\vec{v}_h$ and $\vec{v}_o$ denote the linear velocity of the hand and object.
When the hand has no contact with the object, $r_p$ promotes the hand to move in the same direction as the object. This motion will reduce the impact when the object contacts the hand.
The term $e^{-\norm{p^h-p^o}}$ is a distance weight, motivating the soft catch motion only when the object is close enough.
When the object is in contact with the hand, $r_p$ will reward the hand for staying still to eliminate any undesired arm motion when the object is grasped.
The grasping policy network has the same structure as the reaching policy network, and is also trained with 4 million time steps.

\begin{figure}[t]
    \centering
    \includegraphics[trim=0 270 0 270,clip,width=\linewidth]{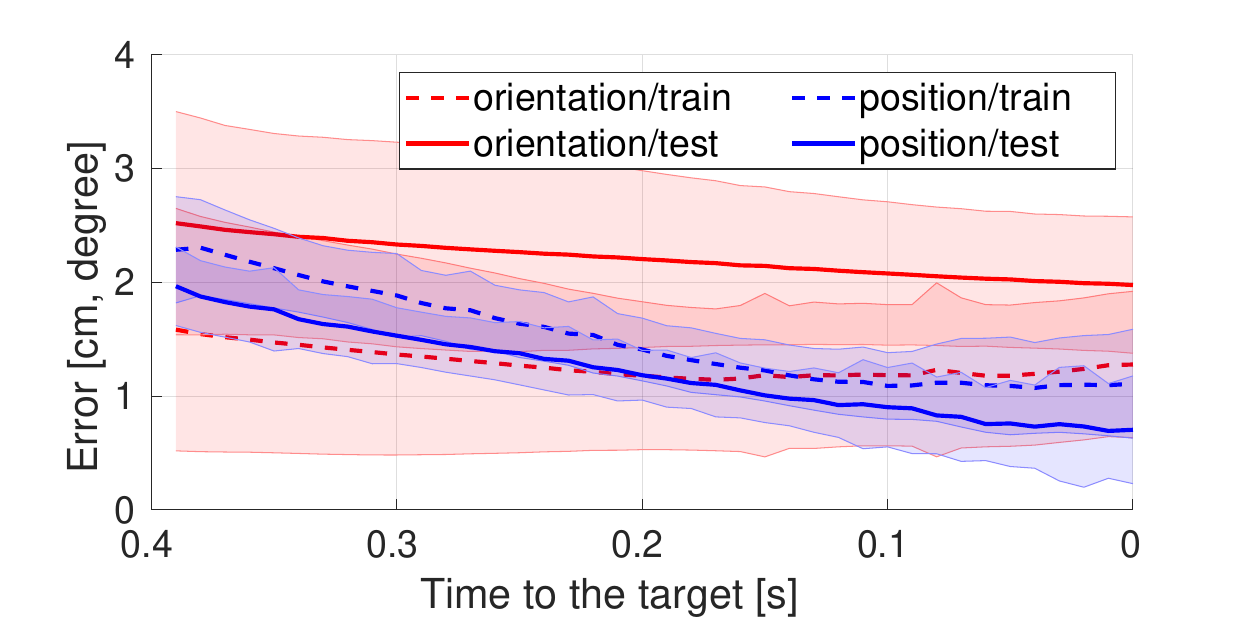}
    \vspace*{-7mm}
    \caption{Decreasing errors of object pose prediction as time elapses. The horizontal axis denotes the remaining time to the target, and the vertical axis is the error between predicted object pose and the ground truth. Solid and dashed lines refer to testing and training objects respectively.}
    \label{fig:prediction_error}
    \vspace{-4mm}
\end{figure}

\subsection{Gating Network} \label{subseq:gate_network}
The reaching policy and the grasping policy play different roles in the catching task, and control the robot at different stages.
Instead of switching the control with hard-coded criteria (e.g. the distance between hand and object), a gating network \cite{MoE_Hinton} is trained with PPO to synthesise the control commands generated from two policies.
The state space of the gating network is the union set of the state spaces of two policies: $S_r \cup S_g$, and it outputs the action weight for the grasping policy $w_g \in [0, 1]$, and the corresponding weight for the reaching policy is $w_r = 1 - w_g$.
The resultant robot action is blended from both policies: $w_g a_g + w_r a_r $, and the reward function is also the weighted sum of the rewards of both policies: $w_r R_r + w_g R_g$. To maximize the reward which is directly related to the robot action, the gating network ought to smoothly switch between two policies at the right time.
With the trained all four aforementioned modules, the gating network is trained in full catching scenarios. The network has the same structure as the reaching and grasping policy network, and also trained with 4 million time-steps.

\section{Validation and Evaluation} \label{seq:validation}
In this section, we first evaluate the performance of each learned module on their corresponding tasks.
Then, the integrated system is validated with catching tasks using multiple novel testing objects.
Though the modules are evaluated in simulation, we mitigate the sim-to-real gap during the training. We introduce model-based air drag into the environment and regulate the robot joint positions, velocities and torques. 
We also add uncertainties to the object pose to simulate vision sensing noises.

\subsection{Validation of each module} \label{subseq:module_validation}
\textbf{Object state prediction network.}
To train the LSTM network, 10,000 flying trajectories of the training objects (as shown in \cref{fig:objects}) with random starting poses and velocities are recorded. Trajectories are split as $90\%$ for training and $10\%$ for testing. 
The object poses are recorded at 100 Hz, and the according velocities and accelerations are computed by numerical differential.
\cref{fig:prediction_error} demonstrates the prediction performance of our LSTM model on the testing dataset. It suggests that the learned model can provide reliable prediction of the object trajectory for the other modules.

\begin{figure}[t]
	\centering
	\subfloat{
    \includegraphics[trim=240 150 240 180, clip, width=0.45\linewidth]{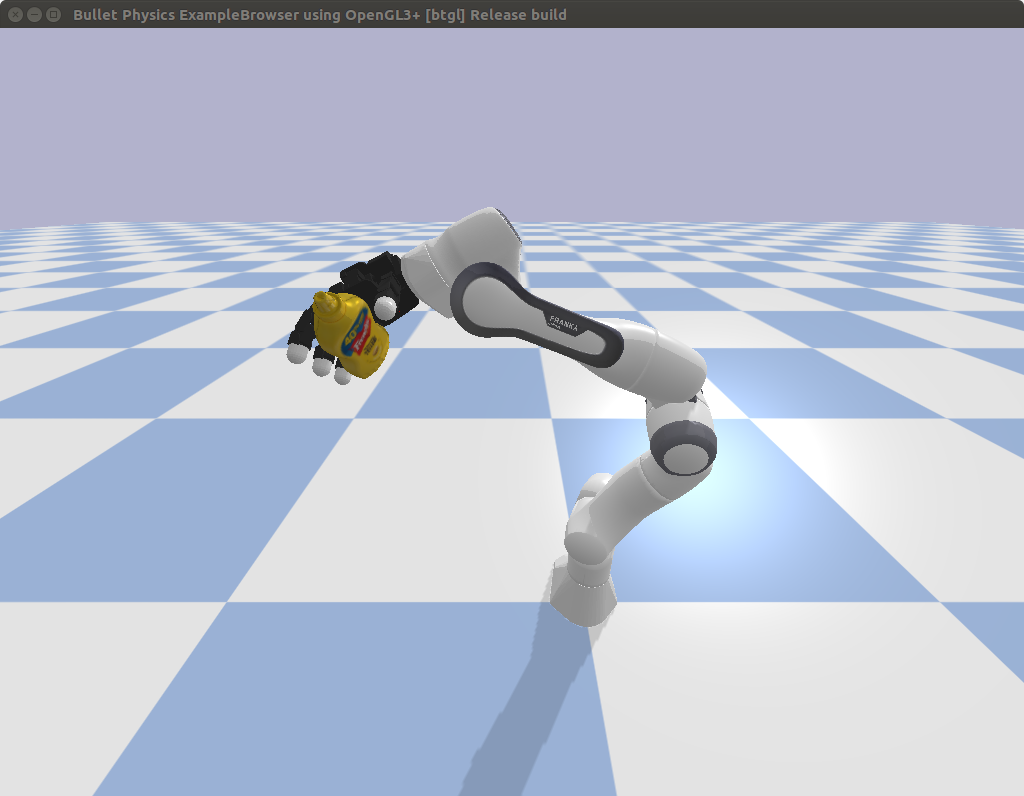}
	}
	\subfloat{
	\includegraphics[trim=240 150 240 180, clip, width=0.45\linewidth]{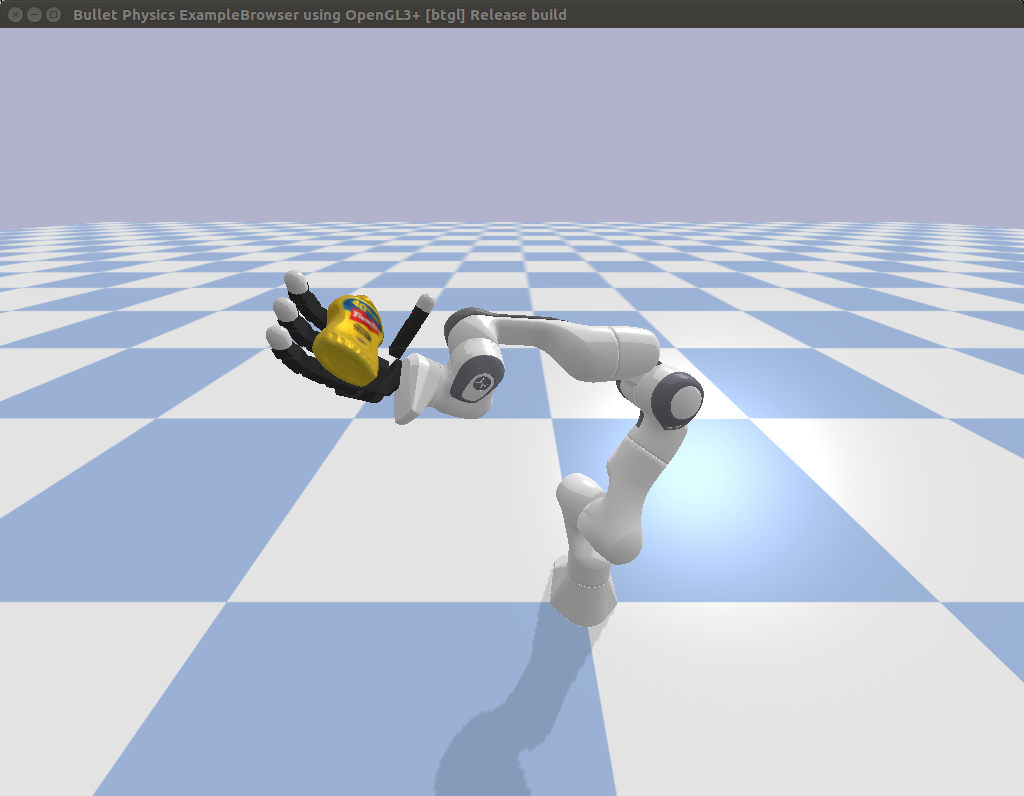}
	}
	\caption{Illustration of how the reaching policy controls the robot to approach the pre-catch pose, based on the pose and moving direction of the object.}
	\label{fig:reaching_policy}
	\vspace{-4mm}
\end{figure}

\textbf{Reaching policy network.} 
For an object fixed at a random pose within the workspace of the robot, the learned reaching policy is capable of moving the robot hand to the proper pre-catch pose as soon as possible.
As shown in \cref{fig:reaching_policy}, for the cylindrical object, the policy can approach the object from the lateral direction, and align the robot hand with the object's major axis.
For the spherical objects, or objects without major axis, the policy can approach them from any direction given by the approaching vector, as long as it is reachable by the robot.

\begin{figure}[t]
    \centering
    \includegraphics[trim=200 150 200 110,clip,width=\linewidth]{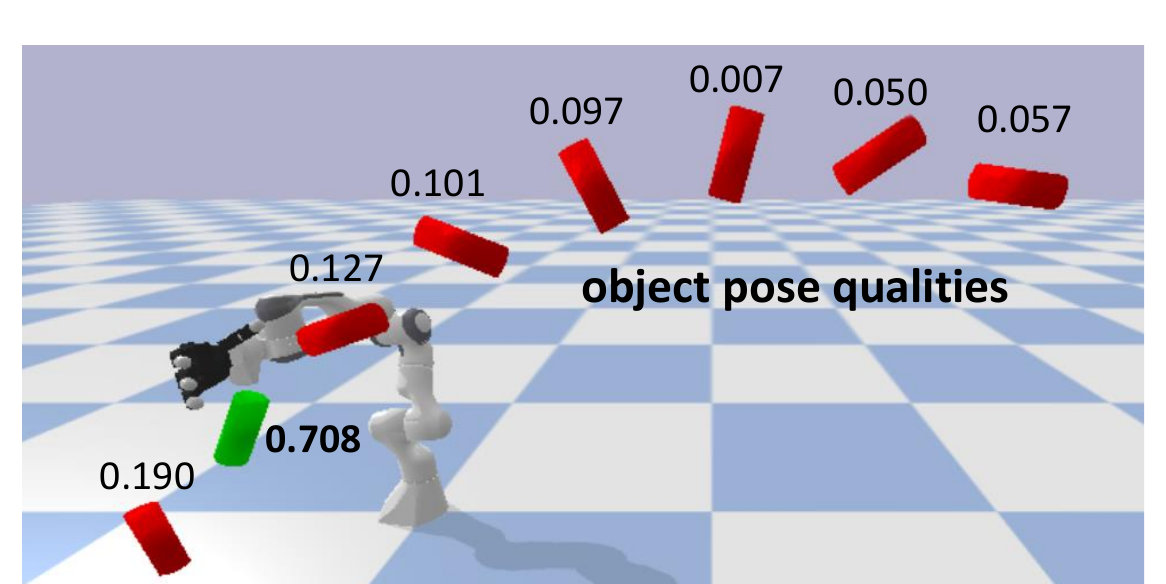}
    \vspace*{-5mm}
    \caption{A representative case of the catching-pose scores (from the quality network) and various object poses. The pose in green is a good example of a high score selected as the target pose for the reaching policy.}
    \label{fig:catch_pose_quality}
    \vspace{-2mm}
\end{figure}

\textbf{Catching pose quality network.}
Taking the target object pose and current hand pose as input, the learned catching pose quality network outputs a score, assessing the effectiveness of the robot to catch the object in the specific condition.
As shown in \cref{fig:catch_pose_quality}, by evaluating every object pose on the predicted trajectory, we can select the one with highest score, as the target for the reaching policy to reach.
Empirically, selecting the object pose with higher score as the catching target is more likely to success, because it is easier for the robot to approach the according adequate pre-catch pose.

\begin{figure}[t]
    \centering
    \includegraphics[trim=300 165 350 200,clip,width=0.8\linewidth]{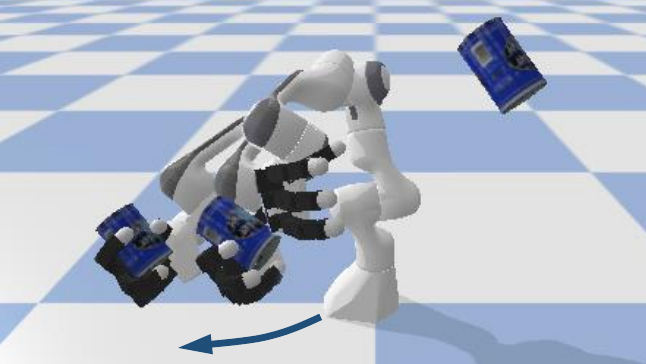}
    \caption{Active reactions of buffering impact during catching. The robot hand moves backward before and after the initial contact with the object and finally holds the object still.}
    \label{fig:soft_catch_motion}
    \vspace{-6mm}
\end{figure}

\begin{figure}[t]
    \centering
    \includegraphics[trim=15 275 15 280,clip,width=\linewidth]{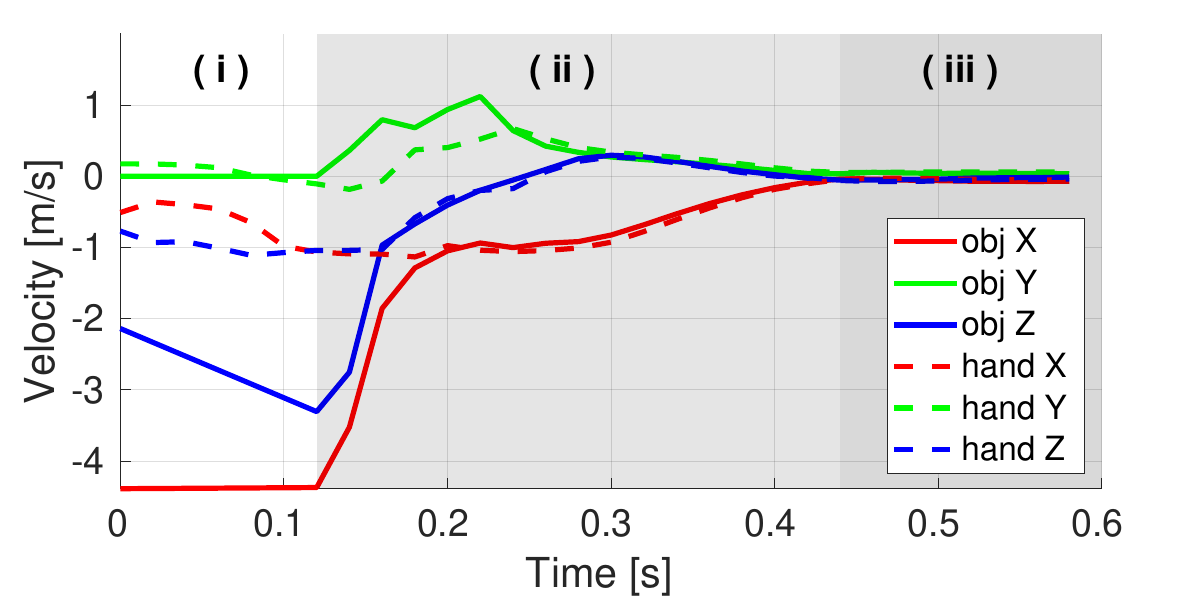}
    \vspace*{-5mm}
    \caption{Linear velocities of the hand and object when the object is close to the robot. Empirically the process can be divided into three phases: (i) pre-catch phase, (ii) catching phase and (iii) holding phase.}
    \label{fig:hand_object_velocity}
    \vspace{-5mm}
\end{figure}

\textbf{Grasping policy network.}
When the hand is close to the pre-catch pose, as the object approaches, the trained grasping policy performs the coordinated motion of the arm and fingers to catch the object with a high speed.
\cref{fig:soft_catch_motion} demonstrates the snapshots of one soft catching motion, and \cref{fig:hand_object_velocity} shows the corresponding velocities of hand and object, from which the motion can be divided into three phases: pre-catch phase, catching phase and the holding phase.
In the pre-catch phase, the hand starts to move along with the object's moving direction, maximizing its translational velocity to reduce the contact impact with the object.
In the catching phase, after the first contact between hand and object, the hand retains movement for a short period of time and then starts to decelerate.
The grasping motion of the fingers is mainly completed in the catching phase.
The soft catching motion provides more time for the fingers to close, and alleviates the bouncing of the object.
In the holding phase, after the grasp is secured, the hand velocities converge to zero and the robot stays still.
The reward term in \cref{eq:reward_velvec} penalizes unnecessary motion after the object is grasped.

\begin{figure}[t]
    \centering
    \includegraphics[trim=0 315 0 310,clip,width=\linewidth]{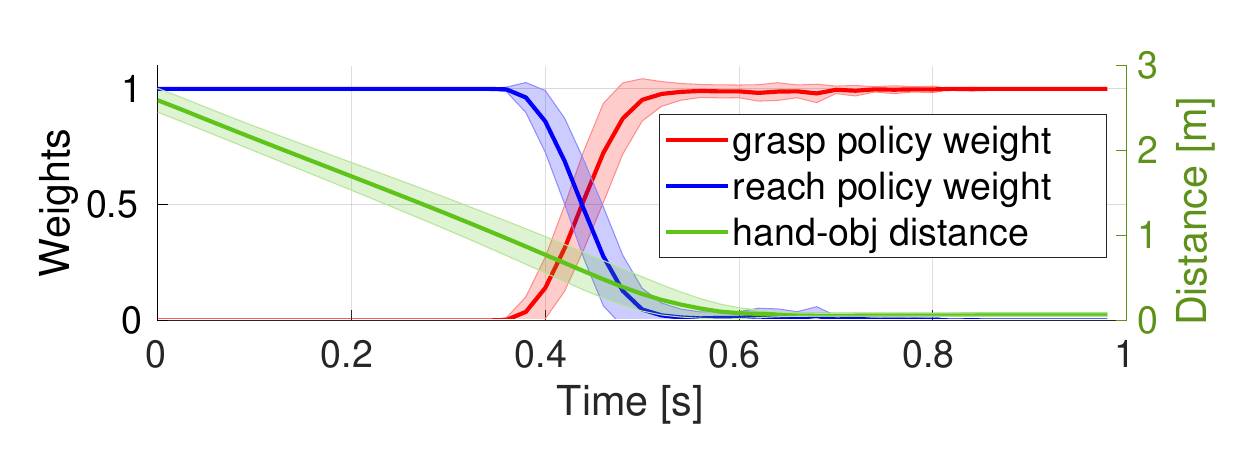}
    \vspace*{-7mm}
    \caption{Action weights for grasping and reaching policies in multiple catching trials, generated by the gating network, and the corresponding hand-object distance. The switch finishes within 0.15 seconds on average.}
    \label{fig:gating_network}
    \vspace{-5mm}
\end{figure}

\textbf{Gating Network.} The learned gating network synthesizes a blend of actions from both control policies by outputting a linear mixture weight. 
As shown in \cref{fig:gating_network}, the reaching policy controls the robot at the beginning. 
The switch starts when there is still some distance between hand and object, leaving enough time for the hand to accelerate to a proper velocity for soft catch. On average, the switch lasts for 0.15 seconds and then the grasping policy takes control for the rest of the time.
To evaluate the effectiveness of gating network, we compare the performance of the integrated catching system with and without gating network.
The hand-coded switch is: if the object is to arrive at the selected target pose after time $T$, the grasping policy controls the robot. Otherwise the reaching policy controls the robot.
Based on the results demonstrated in the first three lines of \cref{tab:grasp_quality}, the learned gating network outperforms the manually designed switch criteria on catching both training and testing objects.

\subsection{Validation of the integrated system} \label{subseq:system_validation}
The integrated system is evaluated by catching various objects in different scenarios. In the catching trials, the robot arm is fixed at the origin and starting from the same configuration. 
Considering the reachability and the max joint velocities of the robot arm, we randomize the object initial position between $[2.75\pm0.25, -0.50\pm0.10, 0.80\pm0.10]$m, orientation in Euler angle between $[0\pm\frac{\pi}{4}, 0\pm\pi, 0\pm\frac{\pi}{4}]$rad, linear velocity between $[-4.5\pm0.5, 0, 3.0\pm0.5]$m/s, and angular velocity between $[0\pm2, 0\pm10, 0]$rad/s.

\begin{table*}[t]
	\centering
    \caption{Success rates of catching various objects in different scenarios, each stemming from a total of 100 trials.}
    \label{tab:grasp_quality}	  
	\def\arraystretch{1.3}
    \begin{threeparttable}
	\begin{tabular}{p{2.66cm} >{\centering}p{0.35cm} | >{\centering}p{0.35cm} >{\centering}p{0.35cm} >{\centering}p{0.35cm}  >{\centering}p{0.35cm} | >{\centering}p{0.35cm} >{\centering}p{0.35cm} >{\centering}p{0.35cm} >{\centering}p{0.35cm} >{\centering}p{0.35cm} >{\centering}p{0.35cm} >{\centering}p{0.35cm} >{\centering}p{0.35cm} >{\centering}p{0.35cm} >{\centering}p{0.35cm} >{\centering}p{0.35cm} >{\centering}p{0.35cm} >{\centering}p{0.35cm} >{\centering}p{0.35cm}}
		\hline
		& & \multicolumn{4}{c}{Training Objects\tnote{\textdagger} [\%]} & \multicolumn{14}{c}{Testing Objects\tnote{\textdagger} [\%]}\tabularnewline
		\cline{2-20}
		& avg. & cld0 & cld1 & cld2 & ball & ybot. & bcan. & rcan. & bana. & wbot. & wood. & chip. & mug & cube & org. & peac. & pear & lemn. & app. \tabularnewline
		\hline
        Integrated system  &80 &75	&73	&72	&97 &81	&74	&78	&66	&71	&81	&72	&83	&90	&85	&86	&82	&86	&85 \tabularnewline
		$T^{*} \in [0.05, 0.20]$ s  &69 &67 &73 &75	&75 &62	&68	&57	&45	&76	&68	&66	&64	&76	&80	&72	&78	&70	&78 \tabularnewline
		$T^{*} = 0.15$ s  &72 &60 &72 &66 &90 &63 &61 &62 &69 &78 &64 &64 &58 &73 &84 &89 &80 &79 &86 \tabularnewline
        $\sigma_{\text{noise}} = [5\text{cm}, 5^{\circ}]$ &64 &60 &63 &63 &73 &48 &56 &52 &38 &54 &59 &56 &64 &77 &75 &72 &77 &77 &84 \tabularnewline
        $\sigma_{\text{noise}} = [15\text{cm}, 15^{\circ}]$ &37 &29 &40 &42 &38 &31 &35 &29 &23 &39 &31 &29 &27 &50 &47 &43 &44 &43 &48  \tabularnewline
        Random perturbation &62 &41	&60	&61	&73 &55 &60	&56	&50	&66	&38	&51	&52	&73	&74	&77	&76	&75	&81 \tabularnewline
		\hline 
	\end{tabular}
	\begin{tablenotes}
      \small
      \item[*] Without using the gating network. $T$ denotes the preparation time for grasping policy, as described in \cref{subseq:grasping_policy}.
      \item[\textdagger] Object list: cylinder0, cylinder1, cylinder2, ball, yellow bottle, blue can, red can, banana, white bottle, wood block, chips can, mug, rubik's cube, orange, peach, pear, lemon, apple. 
    \end{tablenotes}
    \end{threeparttable}
	\vspace{-3mm}
\end{table*}

\begin{figure*}[t]
    \centering
    \includegraphics[width=\linewidth]{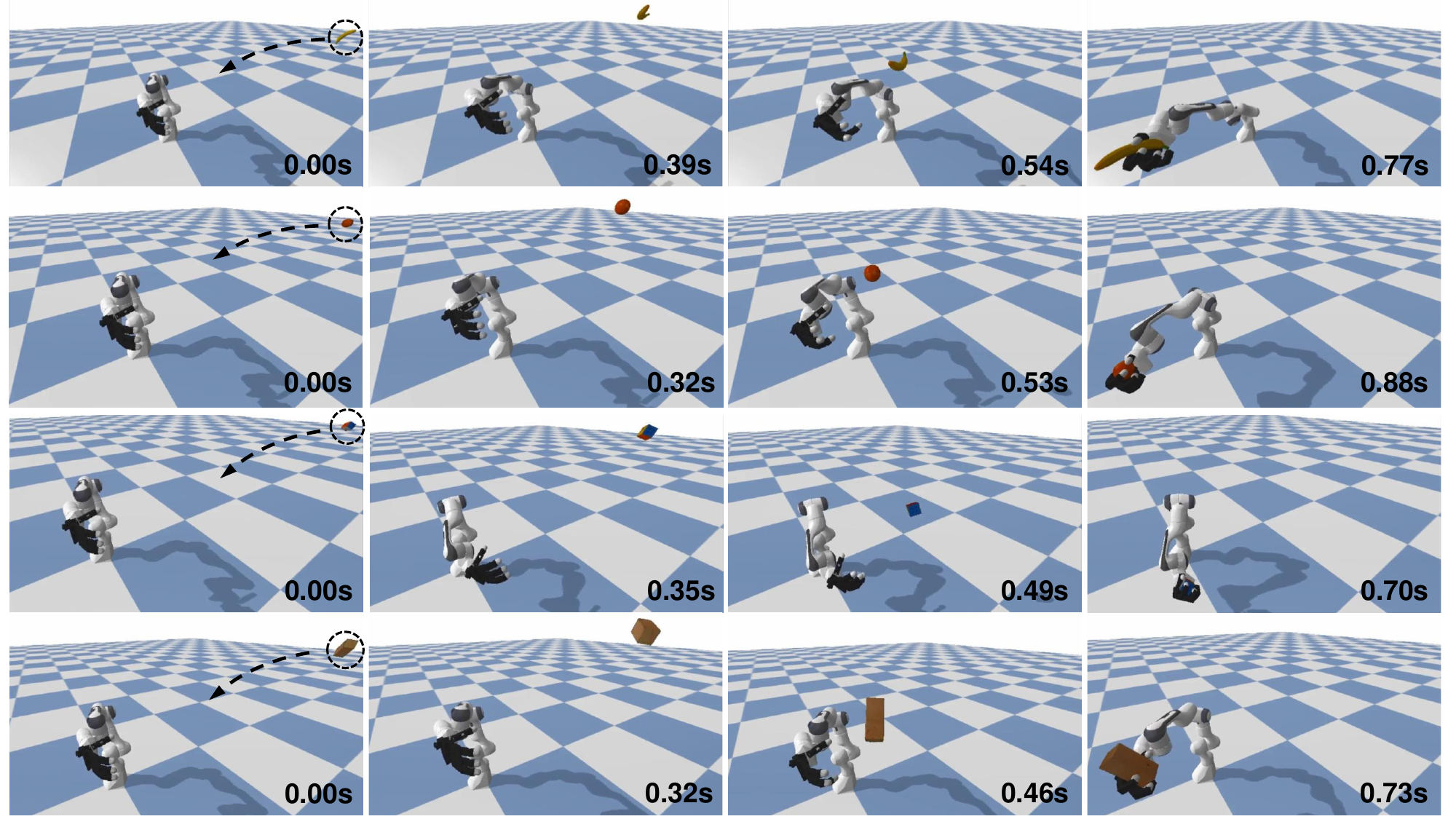}
    \vspace*{-5mm}
    \caption{Validation of the integrated system by catching four new testing objects: a banana, an orange, a Rubik's cube, and a wood block. Initial poses and velocities of objects are randomized within a range.}
    \label{fig:catch_demos}
    \vspace*{-6mm}
\end{figure*}

\textbf{Catching Training Objects}. We first evaluate the catching performance of the integrated system using the four training objects. From \cref{tab:grasp_quality}, the integrated system performs better in catching the sphere than catching the cylinders with different diameters.
For objects with major axes, such as cylinders, it is more challenging to get to the proper pre-catch pose in time, especially when the objects are spinning at high speed.
In contrast, for spherical objects, the robot hand merely needs to move to the nearest pre-catch poses which interact with the flying trajectories, and point the palm against the objects' moving directions, without considering the orientation of the objects. 

\textbf{Catching Testing Objects}. We then perform tests using objects never used during the training. As shown in \cref{fig:objects}, a set of household objects with various shapes and sizes is used to evaluate the generalization ability of our integrated system.
Object models are from YCB dataset \cite{ycb_dataset}.
For simplicity, the objects are rigid with uniformly distributed mass and have the same weight $m=0.3$ kg.
\cref{fig:catch_demos} shows the snapshots of catching four testing objects with different shapes. The robot can reach the pre-catch pose and implement the soft catch motion within 1 second.
As demonstrated in \cref{tab:grasp_quality}, the success rates on catching training and testing objects are comparable, indicating that the learned modules and the integrated system have adequate generalization ability to various sizes and shapes of target objects.

\textbf{Robustness to noise and perturbations}. Furthermore, we test the robustness of our system in the presence of noise in object pose observation, and by introducing a random perturbation to the object during flight.
Additional measurement noise is added to the object position and orientation (Euler angles), as random three-dimensional vectors. The norms of the vectors are sampled from zero-mean Gaussian distributions with standard deviations $\sigma_{\text{noise}}$. To replicate the signal process of noise removal that introduces latency, we process the noisy object pose observations by a low-pass second-order Butterworth filter with a cutoff frequency of 20 Hz, before feed it into the trained modules.
Compared with noise-free state observation, the catching performance with noisy observations drops by $16\%$ and $43\%$ with noise sets $\sigma_{\text{noise}}=[5\text{cm}, 5^{\circ}]$ and $\sigma_{\text{noise}}=[15\text{cm}, 15^{\circ}]$ respectively, indicating the integrated system is robust to sensory noise to some extent. These findings provide a guideline for the requirements of the state-estimation system in future implementations on the real robot. 

To evaluate the system's reaction to sudden changes, a velocity perturbation is applied to the object at a direction of $[1, 0, 1]$ and a magnitude ranging from 0.5 m/s to 1.5 m/s, after the object flies for 0.1 s to 0.2 s. As demonstrated in \cref{fig:first_page_figure} and \cref{tab:grasp_quality}, in most cases the robot can react quickly to the perturbations and catches the object at the new pose.

\begin{figure}[t]
	\centering
	\subfloat[Collision]{
	\includegraphics[trim=0 0 0 0, clip, height=25mm]{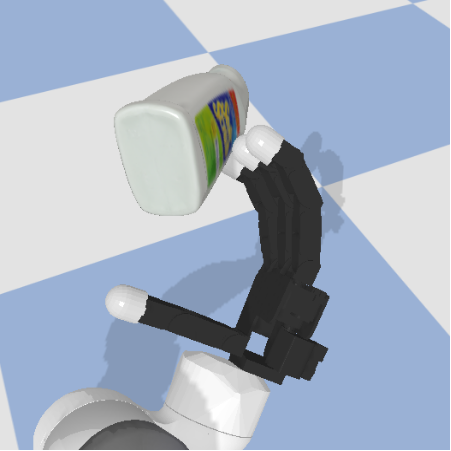}
	\label{subfig:failure_case0}
	}
	\subfloat[Slipping]{
	\includegraphics[trim=0 0 0 0, clip, height=25mm]{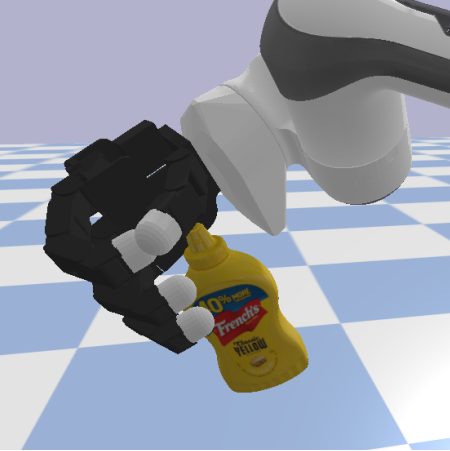}
	\label{subfig:failure_case1}
	}
	\subfloat[Imbalance]{
	\includegraphics[trim=0 0 0 0, clip, height=25mm]{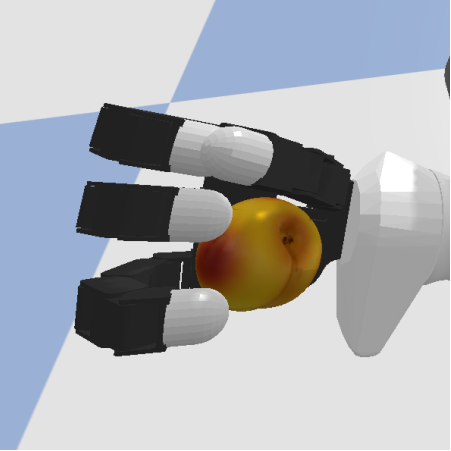}
	\label{subfig:failure_case2}}
	\caption{Representative examples of failure modes: (a) collision before catching, (b) slipping from the grasp or (c) imbalanced contact points.}
	\label{fig:failure_cases}
	\vspace{-5mm}
\end{figure}

\subsection{Failure modes} \label{subseq:failure_cases}
The most common cause for catching failures is the collision before grasping, e.g. between the object and fingertips or the side of the palm. The smallest misalignment of the hand's pre-catch pose can lead to undesired contacts, and therefore this type of failure is more frequent when catching large rotating cylindrical objects, as shown in \cref{subfig:failure_case0}.
Another cause for catching failures is that the object slips from the hand, especially when the object is cylindrical, and its linear velocity is parallel to its major axis, as shown in \cref{subfig:failure_case1}.
For small objects, balanced contact points are important for grasp quality. As shown in \cref{subfig:failure_case2}, the thumb does not contact the object, leading the object to be pushed towards the wrist, and falling out of the hand eventually.

\section{Conclusion} \label{seq:conclusion}
In this work, we proposed a modular learning framework for catching objects in flight. We presented each module and its integration within the system. We employed supervised learning and deep reinforcement learning approaches to train these modules. Extensive tests were performed for each module and the integrated system, including tests with household objects that are never used during training. Furthermore, we studied the robustness and generalization capabilities by performing tests with noisy observations and random perturbations to the objects in flight. 

One potential direction is to reduce the number of modules, e.g. merging the grasping network and gating network.
Sim-to-real transfer remains to be completed as the future work, and our proposed framework needs to be expanded to enable sample-efficient learning from real-world trials.
Currently, only the position and orientation of the object are passed as observations. As part of our future work, we aim to improve catching performance by including the approximate shape of the object into the observation space.
Non-prehensile catching is another interesting and challenging task. For large objects that cannot be grasped by one hand, more dynamic manoeuvres are required by the robot arm to stop and capture them. Developing a unified controller for both prehensile and non-prehensile robot catching can be a promising extension.


\bibliographystyle{IEEEtran}
\balance
\bibliography{reference}
\end{document}